\documentclass[runningheads]{llncs}
\usepackage{graphicx}
\usepackage{cite}
\usepackage{amsmath,amssymb,amsfonts}
\usepackage[ruled,vlined]{algorithm2e}
\usepackage{algorithmic}
\usepackage{efbox,graphicx}
\usepackage{pgfplots}
\pgfplotsset{compat=1.14}
\usepackage{textcomp}
\usepackage{multirow}
\usepackage{soul}
\usepackage{url}

\begin{document}

\title{GPU-based Self-Organizing Maps for Post-Labeled Few-Shot Unsupervised Learning}
\titlerunning{SOM for Post-Labeled Few-Shot Unsupervised Learning}

\author{Lyes Khacef\inst{1} \and
Vincent Gripon\inst{1,2} \and
Beno\^it Miramond\inst{1}}

\authorrunning{L. Khacef et al.}

\institute{Universit\'e C\^ote d'Azur, CNRS, LEAT, France \and
Electronics Dept., IMT Atlantique, France \\
\email{firstname.lastname@univ-cotedazur.fr}}

\maketitle

\begin{abstract}
Few-shot classification is a challenge in machine learning where the goal is to train a classifier using a very limited number of labeled examples. This scenario is likely to occur frequently in real life, for example when data acquisition or labeling is expensive. In this work, we consider the problem of post-labeled few-shot unsupervised learning, a classification task where representations are learned in an unsupervised fashion, to be later labeled using very few annotated examples. We argue that this problem is very likely to occur on the edge, when the embedded device directly acquires the data, and the expert needed to perform labeling cannot be prompted often. To address this problem, we consider an algorithm consisting of the concatenation of transfer learning with clustering using Self-Organizing Maps (SOMs). We introduce a TensorFlow-based implementation to speed-up the process in multi-core CPUs and GPUs. Finally, we demonstrate the effectiveness of the method using standard off-the-shelf few-shot classification benchmarks.


\end{abstract}


\section{Introduction}

In the last decade, Deep Learning (DL) techniques have achieved state-of-the-art performance in many classification problems. However, DL heavily relies on supervised learning with abundant labeled data.
With the fast expansion of Internet of Things (IoT) devices, a huge amount of unlabeled data is gathered everyday, but labeling these data is a very difficult task because of the human annotation cost as well as the scarcity of data in some classes \cite{chen2019look_few_shot}. Finding methods to learn to generalize to new classes with a limited amount of labeled examples for each class is therefore a very active topic of research. This is the main motivation for few-shot learning.
Recently, three main approaches have been proposed in the literature:

\begin{itemize}
    \item \textbf{Hallucination methods} where the aim is to augment the training sets by learning a generator that can create novel data using data-augmentation techniques \cite{chen2019look_few_shot}. However, these methods lack precision which results in coarse and low-quality synthesized data that can sometimes lead to very poor gains in performance \cite{wang2019few_shot_survey}.
    
    \item \textbf{Meta-learning} where the goal is to train an optimizer that initializes the network parameters using a first generic dataset, so that the model can reach good performance with only a few more steps on the new dataset \cite{thrun2012learning_learn}. This type of solution suffers from the domain shift problem \cite{chen2019look_few_shot} as well as the sensitivity of hyper-parameters.
    
    \item \textbf{Transfer learning} where a model developed for a given task is reused as the starting point for a model on a different task. In real-world problems, it happens that we have a classification task in one domain of interest, but we only have sufficient training data in another domain of interest. Therefore, knowledge transfer would greatly improve the performance of learning by avoiding much expensive data-gathering and data-labeling efforts \cite{pan2010transfer_learning}. Hence, transfer learning has emerged as the new learning framework for the few-shot classification task.
\end{itemize}

The problem becomes even harder when facing technical limitations, such as using embedded implementations for real-time processing on the edge. As a matter of fact, in many real-world scenarios, the training data is acquired using the same device that will later be used for training and inference, and labels could be given at any time of the process. To encompass for this added difficulty, we consider in this work the problem of post-labeled few-shot unsupervised learning. In this problem, learning algorithms can be deployed using no annotated data, for example to learn representations using the data acquired by the considered device. These algorithms can later be adjusted using a few labeled samples so that they become able to make predictions, at the condition that this adjustment comes with almost no added complexity to the process, so that it can be performed on the edge.

To address this problem, we propose a solution that combines transfer learning with a recently introduced algorithm~\cite{khacef2019self-organizing_neurons} using Self-Organizing Maps (SOM). On the one hand, transfer learning is used to exploit a Deep Neural Network (DNN) trained on a large collection of labeled data as a ``universal'' feature extractor. On the other hand, a post-labeled clustering algorithm is used to leverage the obtained features and make predictions. This algorithm works in two steps: in a first step, clusters prototypes are learned using no annotated data, then the prototypes are named (labeled) using a few available annotated samples.

The motivation for using the SOM, initially proposed in \cite{kohonen1990som}, comes from the fact they are known to be a very effective clustering method. Indeed, it has been shown that SOMs perform better in representing overlapping structures compared to classical clustering techniques such as partitive clustering or K-means \cite{budayan2009cluster_vs_som}. In addition, SOMs are well suited to hardware implementation based on cellular neuromorphic architectures \cite{sousa2017embedded_som} \cite{khacef2018neuromorphic_hardware} \cite{rodriguez2018grid_som}. Thanks to a fully distributed architecture with local connectivity amongst hardware neurons, the energy-efficiency of the SOM is highly improved since there is no communication between a centralized controller and a shared memory unit, as it is the case in classical Von-Neumann architectures. Moreover, the connectivity and time complexities of the SOM become scalable with respect to the number of neurons \cite{rodriguez2018grid_som}. SOMs are used in a large range of applications \cite{kohonen1996som_app} going from high-dimensional data analysis to more recent developments such as identification of social media trends \cite{silva2018social_media}, incremental change detection \cite{nallaperuma2018bahavior_changes} and energy consumption minimization on sensor networks \cite{kromes2019lorawan}.

This work is an extension of~\cite{khacef2019self-organizing_neurons}, where we used the SOM for MNIST \cite{lecun1998mnist} classification with unsupervised learning, and compared different training and labeling techniques. Here, we focus on the case of few-shot learning, and demonstrate the ability of the proposed method in reaching top performance with the challenging benchmark of mini-ImageNet classification task. We introduce a TensorFlow (TF) software implementation for the proposed method, and compare execution times when using multi-core CPUs or GPUs.

The outline of the paper is as follows. Section \ref{sec_state-of-art} details the SOM training and labeling algorithms and describes the transfer learning methods. Then, Section \ref{sec_methods} presents the mini-ImageNet few-labels classification problem. Next, Section \ref{sec_tf-som} presents the TF-based SOM implementation and shows the multi-core CPU and GPU speed-ups. Afterwards, Section \ref{sec_results} presents the experiments and results on transfer learning with few labels using a SOM classifier. Finally, Section \ref{sec_discussion} and Section \ref{sec_conclusion} discuss and conclude our work.


\section{Proposed methodology}
\label{sec_state-of-art}

In this section, we review the proposed methodology. We begin with the transfer learning part, then how to train the SOM, and we finally explain the labeling procedure.

Let us consider that we are given a dataset $X = \{x, x \in X\}$, that we initially consider to be unlabeled. Our first step consists in extracting relevant features from these inputs.

\subsection{Transfer learning}

In this work, we follow the approach proposed by \cite{hu2020accurate_few_shot} and train a supervised feature extractor $f_{\varphi}$ that we call a \emph{backbone} on a large annotated dataset. The parameters of the backbone are then fixed and used to obtain \emph{generic} features from any input. In our case, we therefore transform $X$ into $V = f_{\varphi}(X) = \{f_{\varphi}(x), x\in X\}$.

\subsection{Self-Organizing Maps learning procedure}

The next step consists in training a SOM using the transformed representations in $V$, i.e. the extracted features. To this end, we use a two-dimensional array of $k$ neurons, that are randomly initialized and updated thanks to the following algorithm, based on the one in~\cite{kohonen1990som}:

\begin{algorithmic}[]
    \STATE
    \STATE \textbf{Initialize} the network as a two-dimensional array of $k$ neurons, where each neuron $n$ with $m$ inputs is defined by a two-dimensional position $p_n$ and a randomly initialized $m$-dimensional weight vector $w_n$.
    \FOR{$t$ from $0$ to $t_f$}
        \FOR{every input vector $v$}
            \FOR{every neuron $n$ in the network}
                \STATE \textbf{Compute} the afferent activity $a_n$ from the distance $d$:
                    \begin{equation}
                    \label{eq_euc-dist}
                        d = \|v - w_n\|
                    \end{equation}
                    \begin{equation}
                    \label{eq_gaussian}
                        a_n = e^{-\frac{d}{\alpha}}
                    \end{equation}
            \ENDFOR
        	\STATE \textbf{Compute} the winner $s$ such that:
            	\begin{equation}
            	\label{eq_max_activity}
                    a_s = \max_{n=0}^{k-1} \left( a_n \right)
                \end{equation}
            \FOR{every neuron $n$ in the network}
            	\STATE \textbf{Compute} the neighborhood function $h_\sigma(t,n,s)$:
                \begin{equation}
                	h_{\sigma}(t,n,s) = e^{-\frac{\|p_n - p_s\|^2}{2{\sigma}(t)^2}}
                \end{equation}
                \STATE \textbf{Update} the weight $w_n$ of the neuron $n$:
                \begin{equation}
                	w_n = w_n + \epsilon(t) \times h_{\sigma}(t,n,s) \times (v - w_n)
                \end{equation}
            \ENDFOR
        \ENDFOR
        \STATE \textbf{Update} the learning rate $\epsilon(t)$:
        \begin{equation}
        	\epsilon(t) = \epsilon_i \left(\frac{\epsilon_f}{\epsilon_i}\right)^{t/t_f}
        \end{equation}
        \STATE \textbf{Update} the width of the neighborhood $\sigma(t)$:
        \begin{equation}
        	\sigma(t) = \sigma_i \left(\frac{\sigma_f}{\sigma_i}\right)^{t/t_f}
        \end{equation}
	\ENDFOR
\end{algorithmic}

It is to note that $t_f$ is the number of epochs, i.e. the number of times the whole training dataset is presented.
The $\alpha$ hyper-parameter is the width of the Gaussian kernel. Its value in Equation \ref{eq_gaussian} is fixed to $1$ in the SOM training, but it does not have any impact in the training phase since it does not change the neuron with the maximum activity. Its value becomes critical though in the labeling process.
The SOM hyper-parameters are reported in Table \ref{tab_hyper-param}.

At the end of the training process, each neuron of the SOM corresponds to a cluster prototype in the considered problem. At this stage, these prototypes are anonymous and cannot be directly used to perform predictions. The next step explains the neurons labeling process for transforming the SOM into a classifier.

\subsection{SOM labeling}
\label{sec_som-labeling}
The labeling is the step between training and test where we assign each neuron the class it represents in the training dataset. We proposed in \cite{khacef2019self-organizing_neurons} a labeling algorithm based on very few labels. The idea is the following: we randomly considered a labeled subset of the training dataset, and we tried to minimize its size while keeping the best classification accuracy. Our study showed that we only need $1 \%$ of randomly taken labeled samples from the training dataset for MNIST classification.
The labeling algorithm detailed in \cite{khacef2019self-organizing_neurons} can be summarized in five steps:

\begin{itemize}
    \item First, we calculate the neurons activations based on the labeled input samples from the euclidean distance following Equation \ref{eq_gaussian}, where $v$ is the input vector, $w_n$ and $a_n$ are respectively the weights vector and the activity of the neuron $n$. The parameter $\alpha$ is the width of the Gaussian kernel that becomes a hyper-parameter for the method.
    
    \item Second, the Best Matching Unit (BMU), i.e. the neuron with the maximum activity is elected.
    
    \item Third, each neuron accumulates its normalized activation (simple division) with respect to the BMU activity in the corresponding class accumulator, and the three steps are repeated for every sample of the labeling subset.

    \item Fourth, each class accumulator is normalized over the number of samples per class.

    \item Fifth and finally, the label of each neuron is chosen according to the class accumulator that has the maximum activity.
\end{itemize}

The complete GPU-based source code for the SOM training, labeling and test is available in \url{https://github.com/lyes-khacef/GPU-SOM}.


\section{Datasets and implementation details}
\label{sec_methods}
\subsection{mini-ImageNet few-shot learning}

In this work, we perform experiments using the mini-ImageNet \cite{vinyals2016imgnet} benchmark. mini-ImageNet is a subset of ImageNet \cite{russakovsky2015imgnet} that contains 60,000 images divided into 100 classes of 600 images, each image has $84\times84$ pixels. Following the standard approach \cite{ravi2017optimization_few_shot}, we use 64 base classes with labels to train the backbone and 20 novel classes to draw the novel datasets from. For each run, 5 classes are drawn uniformly at random among these 20 classes, then $q$ unlabeled inputs and $s$ labelled inputs per class are chosen uniformly at random among the 5 drawn classes. The features of the $(q+s)\times5$ samples are used to train the SOM, then the $s$ labeled samples are used to label the SOM neurons. Finally, the $Q=q\times5$ unlabeled samples are classified and produce a classification accuracy for each run. We run 10,000 random draws to obtain a mean accuracy score and indicate confidence scores (95\%) when relevant.

\subsection{WRN training}

The feature extractor we use is the same as in~\cite{hu2020accurate_few_shot}. It is mostly based on Wide Residual Networks (WRN) \cite{zagoruyko2016wrn} as a backbone extractor, with 28 convolutional layers and a widening factor of 10. As a result, the output feature size (the dimension of a vector $v\in V$) is 640. Let us insist on the fact the backbone is trained on a completely disjoint dataset with the tasks we consider thereafter.

\subsection{Cosine distance}
In transfer learning, the backbone feature extractor is trained with 80 classes that are different from the 20 classes we classify using the SOM. Hence, the features amplitude is not relevant, and the Euclidean distance of the SOM does not provide the best performance. Therefore, we replace the Euclidean distance in Equation \ref{eq_euc-dist} with the Cosine distance in Equation \ref{eq_cos-dist}.

\begin{equation}
\label{eq_cos-dist}
d = 1 - \cos(v, w_n) = 1 - \frac{v . w_n}{\|v\| \times \|w_n\|}
\end{equation}

The Cosine distance is also used in the labeling and test phases. The comparison to Euclidean distance is discussed in Section \ref{sec_discussion}.


\section{SOM software implementation}
\label{sec_tf-som}

\subsection{TensorFlow-based SOM}
The SOM was implemented using TF \cite{abadi2016tensorflow} 2.1, an end-to-end open source platform for machine learning that uses dataflow graphs to represent computation, shared state, and the operations that mutate that state. It maps the nodes of a dataflow graph across multiple computational devices including multi-core CPUs, general-purpose GPUs and custom-designed ASICs known as Tensor Processing Units (TPUs) \cite{abadi2016tensorflow}. TF facilitates the design of many machine learning models providing built-in functionalities such as convolution, pooling and dense (i.e. fully connected) layers.
However, TF does not provide computational neuroscience models, and to the best of our knowledge, there is no efficient implementation for SOMs using TF.

\subsection{CPU and GPU speedups}

\begin{figure}[ht]
	\centerline{\includegraphics[width=1.0\linewidth]{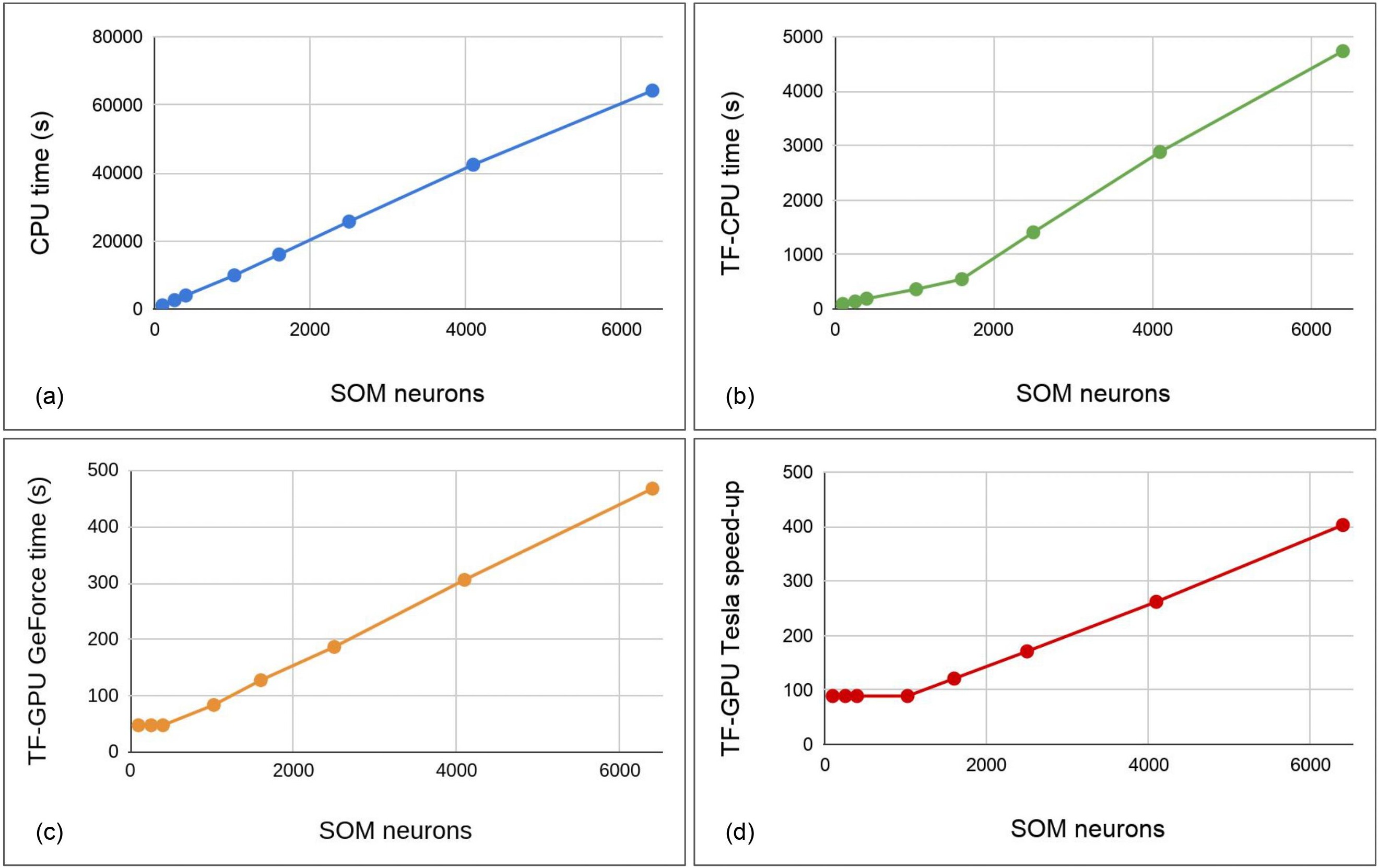}}
	\caption{SOM training speed on MNIST database for 10 epochs (i.e. 600,000 samples of 784 dimensions) VS. number of SOM neurons: (a) CPU (mono-core) implementation; (b) TF-CPU (muti-core) implementation; (c) TF-GPU GeForce implementation; (d) TF-GPU Tesla implementation.}
	\label{fig_som-speed}
\end{figure}

The SOMs of different sizes were trained for 10 epochs on MNIST database, i.e. 600,000 samples of 784 dimensions. The CPU mono-core implementation is based on NumPy \cite{vanderwalt2011numpy} and run on an Intel Core i9-9880H CPU (16 cores), while the GPU implementation is based on TF 2.1 \cite{abadi2016tensorflow} and run on two different GPUs: Nvidia GeForce RTX 2080 and Nvidia Tesla K80 freely available on Google Colab cloud service \cite{carneiro2018colab}. Interestingly, the TF-based SOM can also run on the multiple cores of the CPU, providing a speed-up even without access to GPU.

Figures \ref{fig_som-speed}-a , \ref{fig_som-speed}-b, \ref{fig_som-speed}-c and \ref{fig_som-speed}-d show that the time complexities of the CPU, TF-CPU and TF-GPU implementations are all linear. It is to note that the time complexity slope of the TF-CPU, TF-GPU GeForce and TF-GPU Tesla implementations changes at 1600 neurons, 400 and 1024 neurons respectively, which is due to their different degrees of parallelism.

\begin{figure}[ht]
	\centerline{\efbox{\includegraphics[width=0.7\linewidth]{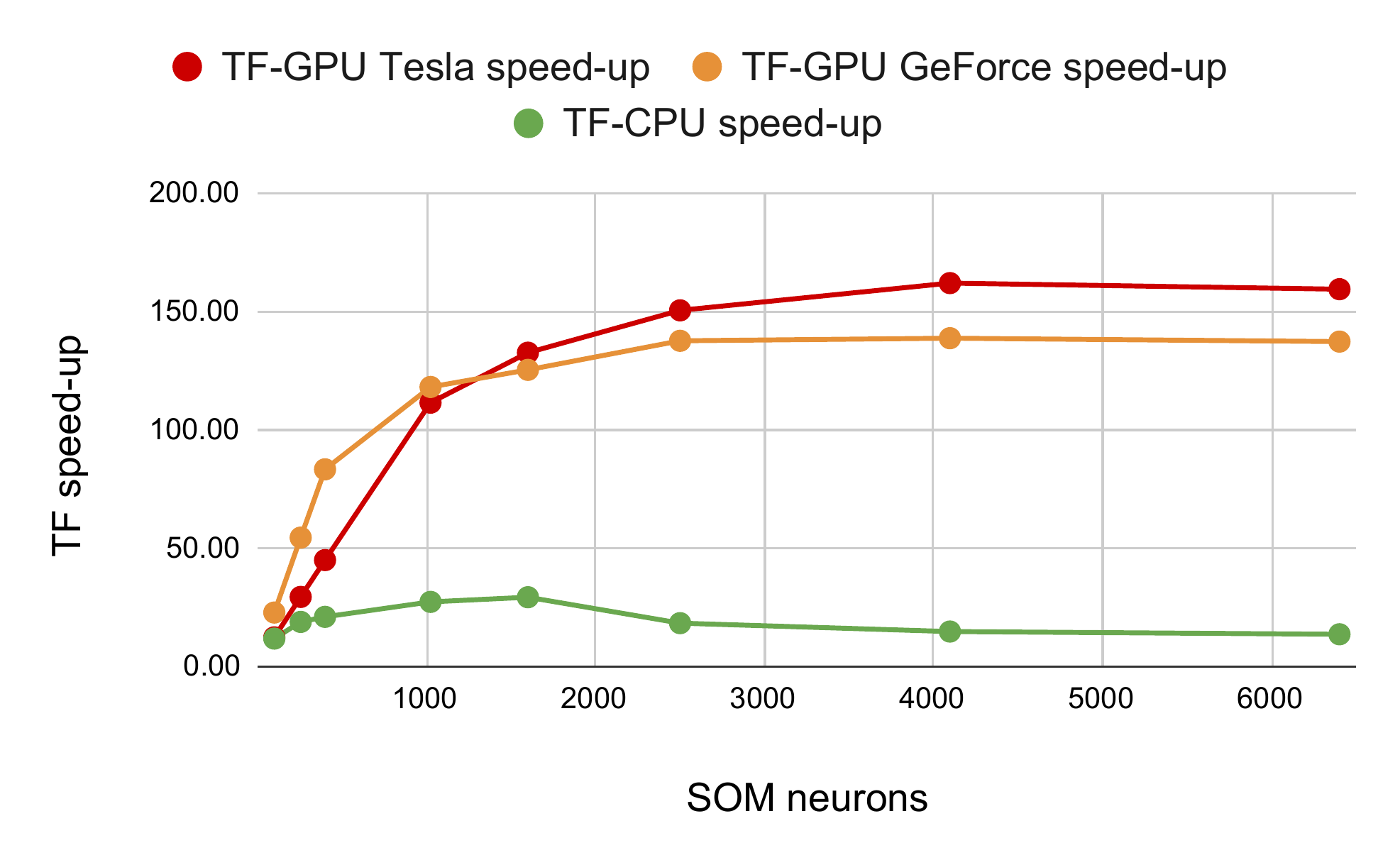}}}
	\caption{TF-CPU and TF-GPU speed-ups compared to CPU.}
	\label{fig_som-speedup}
\end{figure}

As shown in Figure \ref{fig_som-speedup}, we achieved a minimum speedup of $12\times$ ($22\times$) and a maximum speedup of $161\times$ ($138\times$) with the TF-GPU Tesla (TF-GPU GeForce) implementation, with an increasing speedup with respect to the number of neurons. Our GPU implementation is therefore scalable in simulation time with respect to the SOM size, which is an important aspect to accelerate the simulations and hyper-parameters exploration. 
In addition, we achieved a minimum speedup of $11\times$ times and a maximum speedup of $29\times$ times with the TF-CPU implementation, which runs the 16 cores of the CPU. Nevertheless, the gap between the GPU and CPU speed-ups increases with the number of neurons, which is expected due to the highly parallel computation of the GPU hardware.

Recent works have tried an other approach using CUDA acceleration on Nvidia GPUs. They showed relative gains to CPU of $44\times$ \cite{moraes2012som_cuda}, $47\times$ \cite{gavval2019som_cuda} and $67\times$ \cite{mcconnell2012som_opencl_cuda}. Our implementation reaches an average gain of $19\times$ in a multi-core Intel Core i9 CPU, $100\times$ in a Nvidia Tesla GPU and $102\times$ in a Nvidia GeForce GPU. A fair comparison is difficult since we do not target the same hardware, but the order of magnitude is comparable and our results are in the state of the art.
Another advantage of our TF-based approach is the easy integration of the SOM layer into Keras \cite{chollet2015keras}, a high-level neural networks API capable of running on top of TF with a focus on enabling fast experimentation.


\section{Experiments and results}
\label{sec_results}
The SOM training hyper-parameters for the different settings were found with a grid search and are reported in Table \ref{tab_hyper-param}.

\begin{table}[ht]
\centering
\caption{SOM training hyper-parameters.}
\label{tab_hyper-param}
\begin{center}
\resizebox{0.5\linewidth}{!}{
    \begin{tabular}{l c c c c c c}
    \hline
    \textbf{Dataset} & $\epsilon_i$ & $\epsilon_f$ & $\eta_i$ & $\eta_f$ & Epochs & $\alpha$ \\ \hline
    \textbf{mini-ImageNet}    & 1               & 0.01            & 10              & 0.1             & 10              & 1              \\ \hline
    \end{tabular}
}
\end{center}
\end{table}

\begin{figure}[h!]
	\centerline{\efbox{\includegraphics[width=0.7\linewidth]{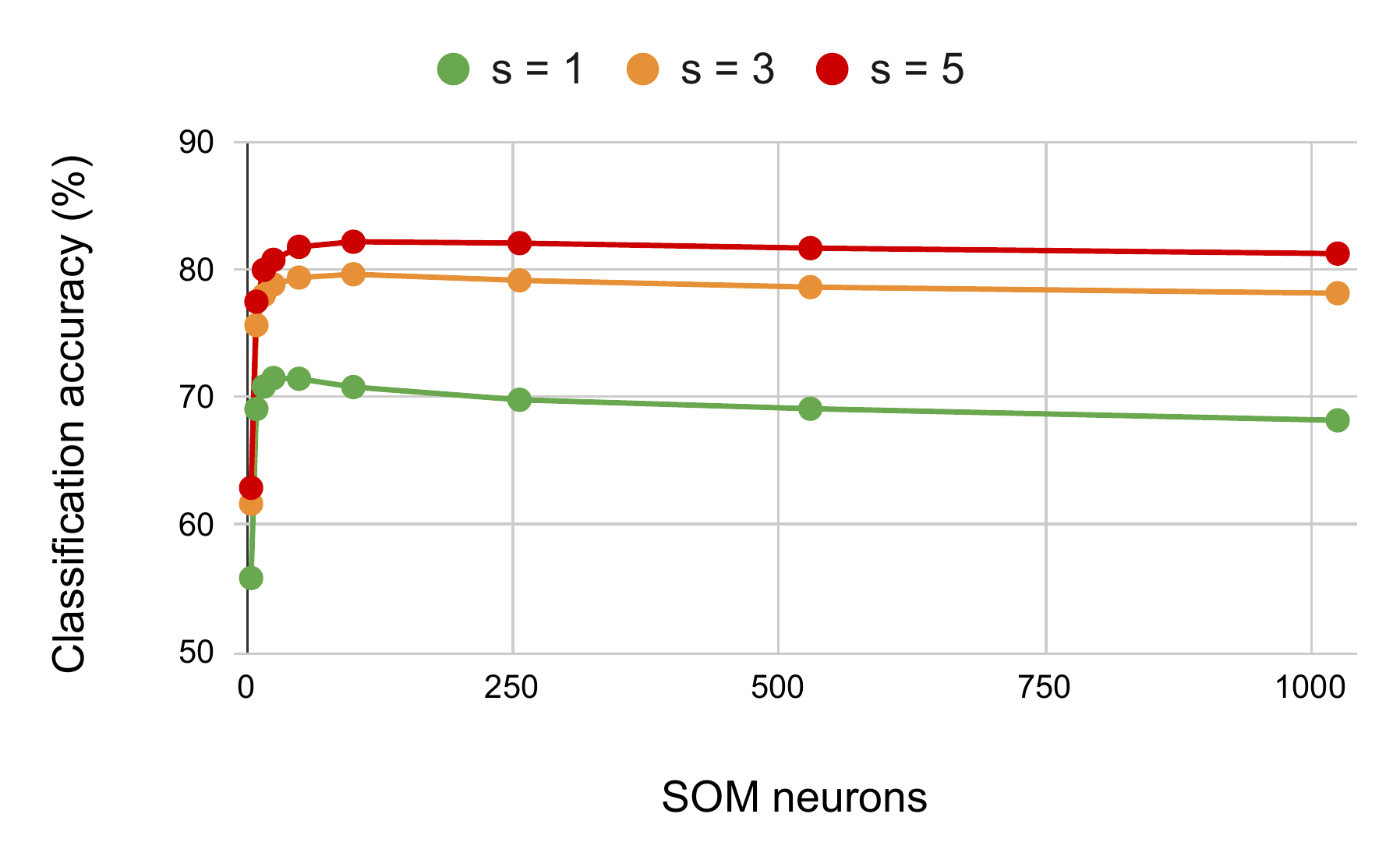}}}
	\caption{SOM classification accuracy on mini-ImageNet transfer learning for different numbers of labeled samples s vs. number of SOM neurons.}
	\label{fig_wrnsom-neurons}
\end{figure}

First, we investigated the impact of the SOM size on the classification accuracy for the commonly used number of unlabeled samples $q = 15$ and labeled samples $s = [1, 3, 5]$ \cite{hu2020accurate_few_shot}. Figure \ref{fig_wrnsom-neurons} shows that there is an optimal point at 25 neurons for $s = 1$ and 100 neurons for $s = 3$ and $s = 5$. There is a tradeoff between the number of neurons that learn different prototypes and the quality of the learning/labeling of these neurons. The more neurons we have, the more potential to learn different prototypes of the data but the more fuzzy the prototypes become, which makes the labeling part more difficult. For example, a neuron may be assigned a class ``A'' with respect to the labeled subset, but will be more active for a class ``B'' with respect to the test set. When we only have one labeled sample per class, i.e. $s = 1$, then a SOM of only 25 neurons achieves the best accuracy because more neurons will not converge as well.

\begin{figure}[h!]
	\centerline{\efbox{\includegraphics[width=0.7\linewidth]{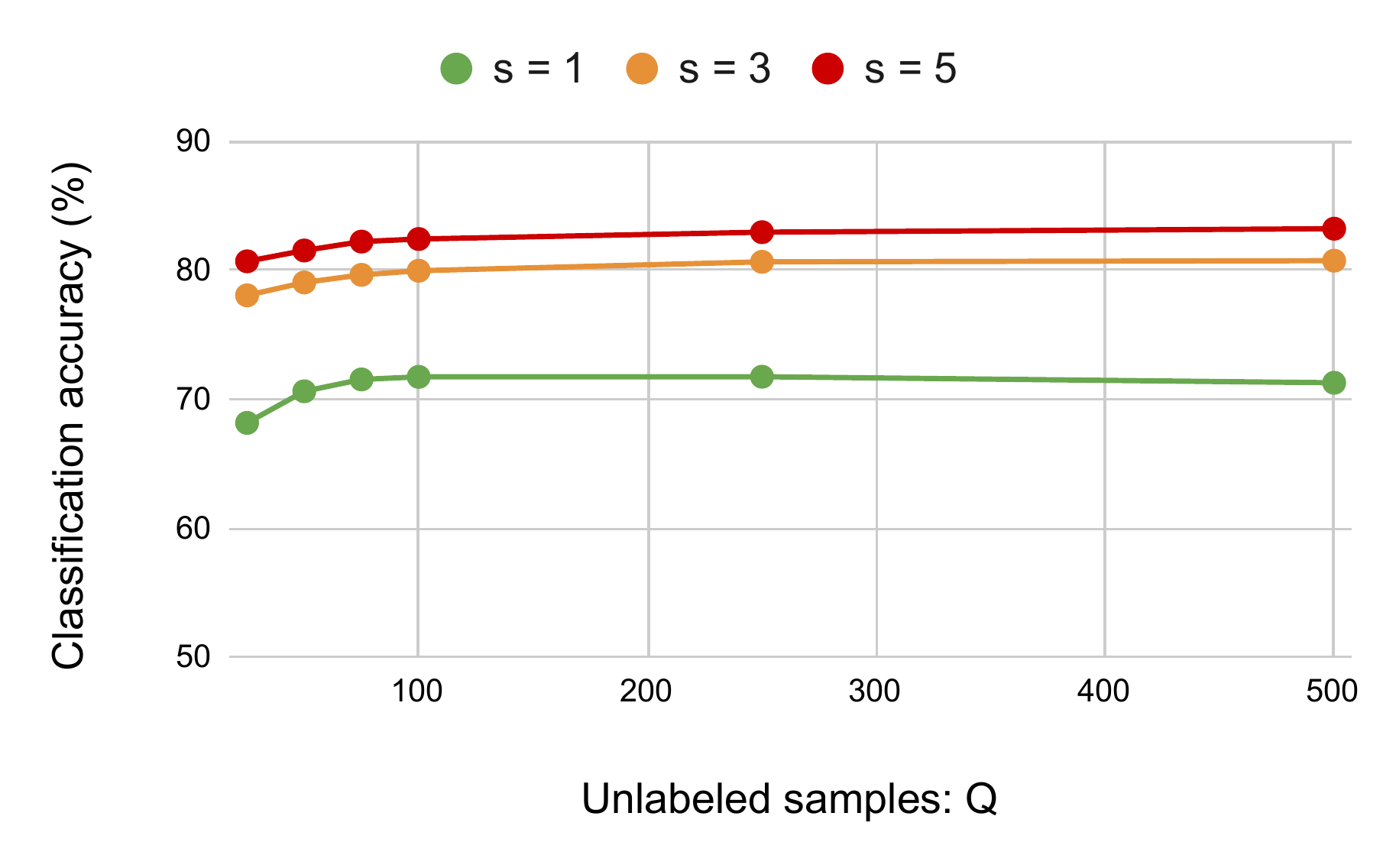}}}
	\caption{SOM classification accuracy on mini-ImageNet transfer learning for different numbers of labeled samples s vs. number of unlabeled samples to classify Q.}
	\label{fig_wrnsom-data}
\end{figure}

Next, we varied the number of unlabeled data $Q = q\times5$ with the above mentioned SOM sizes. Figure \ref{fig_wrnsom-data} shows that even though the labels are only used for the neurons class assignment and not in the training process, they still have a large impact on the accuracy. Naturally, the more labeled data we have, the better accuracy we get. 
A second remark is that the more unlabeled data we have, the better accuracy we get too. This is not intuitive, because the unlabeled data are the queries, i.e. the samples to classify, so the more we have the harder the classification task becomes. However, since the SOM is trained on these data, its adaptation capabilities makes the accuracy increase with the number of unlabeled data for the same number of labels. The only exception is when $s = 1$, where there is a small decrease in accuracy between $Q = 250$ ($71.74\% \pm 0.21$) and $Q = 500$ ($71.27\% \pm 0.21$).
A third remark is that the SOM reaches the same accuracy for $[s = 5, Q = 25]$ and $[s = 3, Q = 250]$, which means that the lack of labeled data can be compensated by more unlabeled data. In fact, it is a very interesting property since unlabeled data can be gathered much more easily, and no extra-effort for labeling these data is needed.


\section{Discussion}
\label{sec_discussion}

\begin{figure}[h]
	\centerline{\efbox{\includegraphics[width=0.7\linewidth]{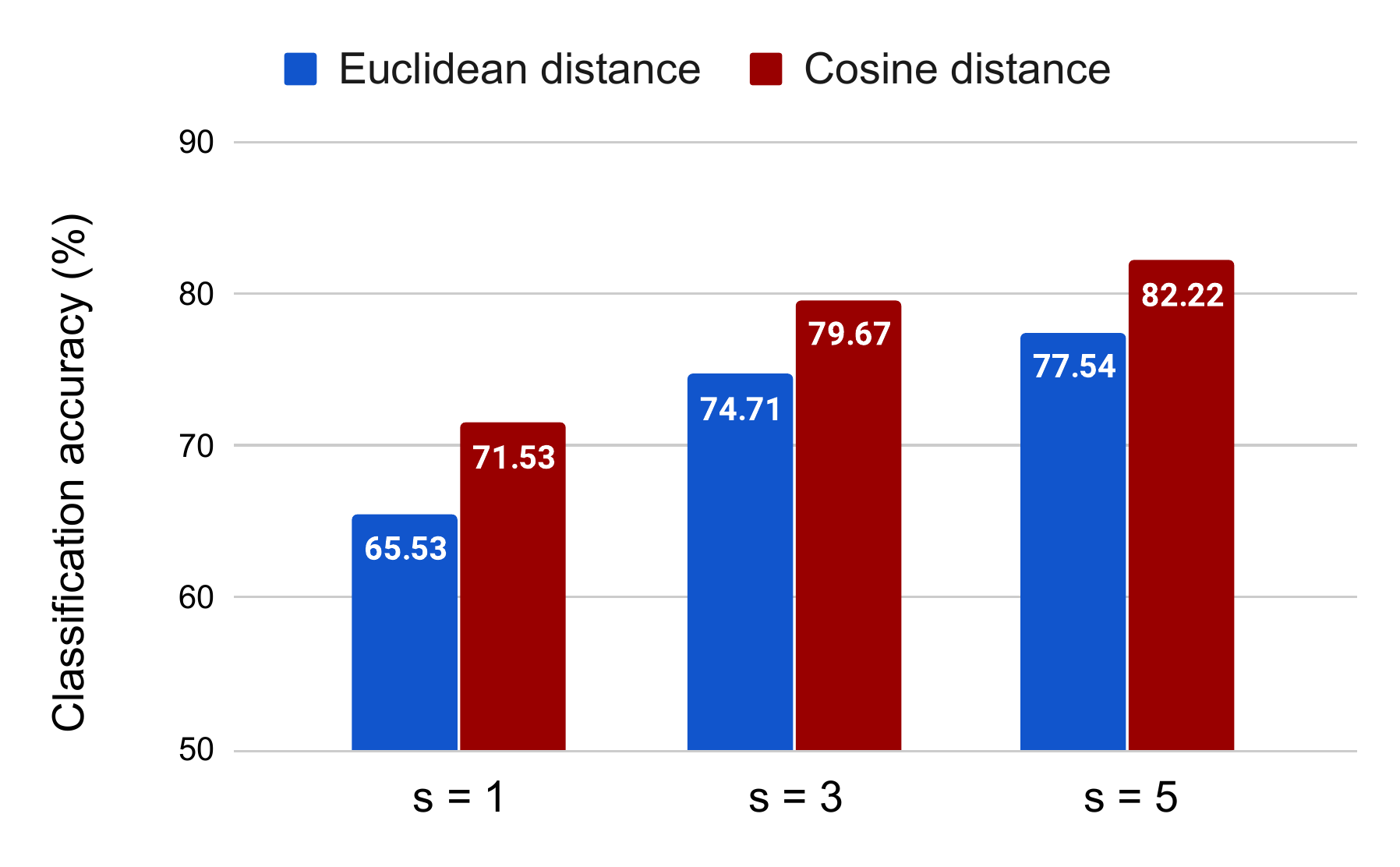}}}
	\caption{SOM classification accuracy on mini-ImageNet transfer learning with few labels using Euclidean distance and Cosine distance.}
	\label{fig_cosine-distance}
\end{figure}

The choice of using the Cosine distance in the SOM computation (training, labeling and test) was inspired from the work of \cite{hu2020accurate_few_shot}. In fact, Figure \ref{fig_cosine-distance} shows that replacing the Euclidean distance by the Cosine distance significantly improves the SOM classification accuracy, with a gain of +5.9\%, +4.96\% and +4.68\% for s = 1, s = 3 and s = 5, respectively. It validates our hypothesis about the non-effectiveness of the Euclidean distance when using transfer learning.

\begin{table}[h]
\centering
\caption{mini-ImageNet few labels transfer learning with q = 15 (Q = 75): state of the art reported from \cite{hu2020accurate_few_shot}.}
\label{tab_imgnet}
\begin{center}
\resizebox{1.0\linewidth}{!}{
    \begin{tabular}{l l l c c}
    \hline
    \textbf{Method}              & \textbf{Backbone} & \textbf{Classifier} & \textbf{1-shot (\%)}  & \textbf{5-shot (\%)}  \\ \hline
    wDAE-GNN \cite{gidaris2019gnn_ae}             & WRN               & Supervised          & 61.07 ± 0.15          & 76.75 ± 0.11          \\
    ACC+Amphibian \cite{snell2017prototype_networks}       & WRN               & Supervised          & 64.21 ± 0.62          & \textbf{87.75 ± 0.73} \\
    BD-CSPN \cite{liu2019prototype_rectification}             & WRN               & Supervised          & 70.31 ± 0.93          & 81.89 ± 0.60          \\
    Transfer+SGC \cite{hu2020accurate_few_shot} & WRN               & Supervised          & \textbf{76.47 ± 0.23} & 85.23 ± 0.13          \\
    Transfer+SOM {[}Our work{]}  & WRN               & \textbf{Unsupervised}        & 71.53 ± 0.23          & 82.22 ± 0.15          \\ \hline
    \end{tabular}
}
\end{center}
\end{table}

Finally, Table \ref{tab_imgnet} reports the recent works that proposed solutions to the mini-ImageNet few labels classification problem using transfer learning with the WRN backbone feature extractor. The SOM reaches top-2 accuracy for $s = 1$ and top-3 accuracy for $s = 5$, which is a good result that proves the SOM ability to handle complex datasets. 
Nevertheless, one has to keep in mind that while the other works use the few labels in the training process, we only use them for neurons labeling phase. Our accuracy performance is therefore obtained with fully unsupervised learning followed by post-labeling, which we believe is the right approach for the few-shot classification problem in the context of embedded systems on the edge.


\section{Conclusion and further works}
\label{sec_conclusion}
We introduced in this work the problem of post-labeled few-shot unsupervised learning and proposed a solution that combines transfer learning and SOMs. Transfer learning was used to exploit a WRN backbone trained on a base dataset as a feature extractor, and the SOM was used to classify the obtained features from the target dataset. The SOM is trained with no label, then labeled with the few available annotated samples. We show that we reach a good performance on the mini-ImageNet few shot classification benchmark with an unsupervised learning method. Furthermore, the SOM is suitable for hardware implementations based on a cellular neuromorphic architecture, which enables its application on the edge. Finally, to speed-up the SOM simulation process, we proposed a novel TF-based GPU implementation which is about $100\times$ faster than the classical CPU implementation.


\section*{Acknowledgment}
This work has been supported by the French government, through the UCAJEDI Investments in the Future project managed by the National Research Agency (ANR) with the reference number ANR-15-IDEX-01.


\end{document}